\documentclass[11pt]{article}

\usepackage[final]{acl}

\usepackage{fontspec} % use this with XeLaTeX instead of times
\usepackage{latexsym}

\usepackage[T1]{fontenc}
\usepackage[utf8]{inputenc}

\usepackage{microtype}

\usepackage{inconsolata}

\usepackage{graphicx}

\usepackage{times}
\usepackage{threeparttable}
\usepackage{booktabs, multirow}
\usepackage{float}
\usepackage{amsmath}
\usepackage{amssymb}

\title{H\textsuperscript{2}Table: Hierarchical Hypergraph-Enhanced Large Language Models for Complex Table Reasoning}

\author{
  \textbf{Jia Ling\textsuperscript{1}},
  \textbf{Yangfan Wang\textsuperscript{1}},
  \textbf{Chen Tang\textsuperscript{2}},
  \textbf{Haoming Tan\textsuperscript{1}},
\\
  \textbf{Yang Yang\textsuperscript{3}},
  \textbf{Yi Guan\textsuperscript{1}},
  \textbf{Jingchi Jiang\textsuperscript{1,4,*}}
\\
  \textsuperscript{1}Harbin Institute of Technology, 
  \textsuperscript{2}AI Research Center, Midea Group (Shanghai) Co., Ltd. \\
  \textsuperscript{3}Changchun University of Science and Technology \\
  \textsuperscript{4}State Key Laboratory of Smart Farm Technologies and Systems\\
  \texttt{\{2022112476,yf.wang,2022111514\}@stu.hit.edu.cn} \\ \texttt{travistang@foxmail.com}, 
  \texttt{yangyang\_hit\_wi@163.com} \\ 
  \texttt{\{guanyi,jiangjingchi\}@hit.edu.cn}
}
\begin{document}
  \maketitle
  {\let\thefootnote\relax\footnotetext{\textsuperscript{*}Corresponding author.}}
  \begin{abstract}
    Tables are ubiquitous across diverse domains, yet reasoning over them remains a significant challenge for modern large language models (LLMs). Current approaches typically linearize tables into sequences, inherently overlooking their intrinsic two-dimensional and hierarchical structure. To address this, we propose \textbf{H\textsuperscript{2}Table} (\textbf{H}ierarchical \textbf{H}ypergraph-Enhanced \textbf{Table} Reasoning), a novel framework that represents complex tables as hierarchical nested hypergraphs. To process this representation, we design a tailored hypergraph encoder to facilitate message passing between hyperedges (headers) and nodes (cells), thereby perceiving the semantic entailment relationships between them within complex tables. Furthermore, we introduce a set of learnable query vectors acting as a lightweight bridge to extract representative structural embeddings from the encoder into the LLM. Experimental results demonstrate that our approach effectively handles complex table question answering tasks with hierarchical nested headers. Notably, on the HiTab dataset, H\textsuperscript{2}Table achieves an average  improvement of 22.88\% over state-of-the-art baselines on highly complex tables with a nesting depth of four. Our code is available at: \url{https://github.com/lila120/h2table}.
  \end{abstract}

  \section{Introduction}

  Table understanding and reasoning are critical for enabling diverse downstream applications, such as table question answering (TableQA) \cite{pasupat2015compositional} and Text-to-SQL \cite{yu2018spider}, to process complex, real-world data. Consequently, leveraging the powerful semantic synthesis and reasoning capabilities of large language models (LLMs) has become essential in this domain. Driven by the need to decode structured data, existing approaches have evolved significantly. Recently, the paradigm has shifted toward LLM-based methods. These approaches primarily rely on 1D text-centric serialization to harness pre-trained knowledge, or construct massive instruction datasets for full-parameter supervised fine-tuning (SFT) (e.g., TableLlama \citealp{zhang2024tablellama}, TableGPT \citealp{li2024table}). Furthermore, to explicitly capture internal structural information, some works represent tables as graphs. For instance, HeGTa \cite{jin2025hegta} models tables as heterogeneous graphs, while HYTREL \cite{chen2023hytrel} utilizes hypergraphs. Building upon this structural perspective, TAMO \cite{li2026table} integrates the hypergraph encoder from HYTREL with LLMs, achieving the current state-of-the-art performance.

  \begin{figure}[t]
  \centering
  \includegraphics[width=0.48\textwidth]{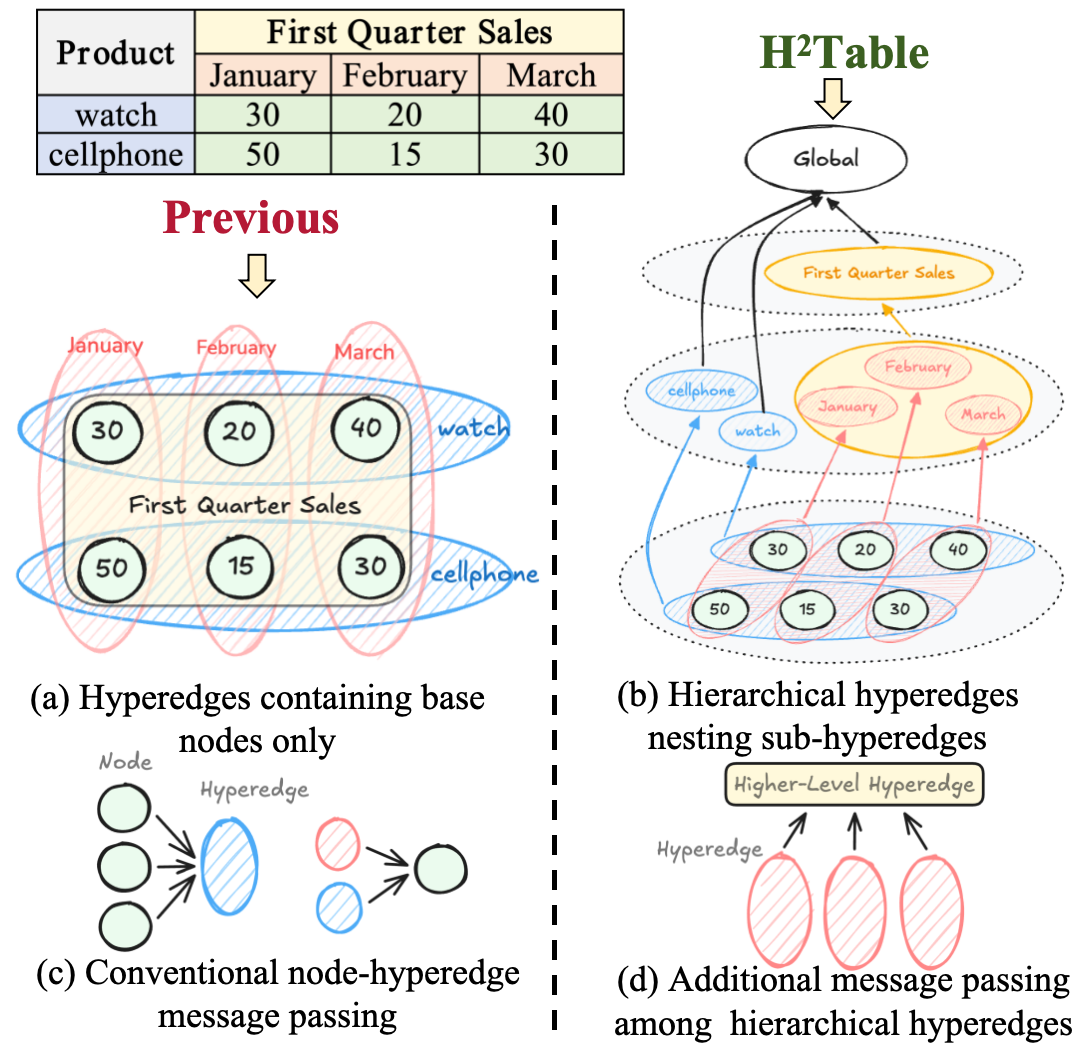}
  \caption{Comparison of modeling approaches and corresponding message passing mechanisms for complex tables between the proposed H\textsuperscript{2}Table and other leading baselines.}
  \label{fig:motivation}
\end{figure}

 However, existing methods exhibit several critical limitations. LLM paradigms relying on 1D serialization and full-parameter fine-tuning demand massive amounts of high-quality data and incur prohibitive computational costs. Alternatively, representing tables as standard graphs fails to capture their intrinsic structural features. Since edges in standard graphs are strictly pairwise, they inherently contradict the one-to-many relationships between headers and cells. Existing hypergraph-based methods like HYTREL overcome this bottleneck but are restricted to simple flat tables. Furthermore, lacking LLM integration, they are incapable of handling generative tasks like TableQA. While TAMO addresses this by coupling a hypergraph encoder with an LLM, its encoding mechanism fundamentally overlooks the nested hierarchy of complex headers. As illustrated in Figures \ref{fig:motivation}(a) and (c), when processing complex tables with hierarchical headers, TAMO flattens all hyperedges into a single level by directly connecting high-level headers to their associated cell nodes. This approach neglects the hierarchical relationships among hyperedges, leading to a loss of multi-level dependencies and table semantic implications. Consequently, the subsequent message passing occurs solely between these flattened hyperedges and cell nodes, ignoring the essential interactions across different header levels.

  To address the aforementioned challenges, we propose H\textsuperscript{2}Table (\textbf{H}ierarchical \textbf{H}ypergraph-Enhanced \textbf{Table} Reasoning). We first introduce a tailored transformation that maps complex tables into hierarchical nested hypergraphs. Rather than flattening these structures, our method preserves intrinsic tabular semantics, particularly the semantic entailment flowing from multi-level headers down to specific cells, by employing higher-level hyperedges that connect parent headers exclusively to their direct sub-headers. To encode these semantic dependencies, we develop a hypergraph encoder equipped with a four-stage hierarchical interaction module (V2E, C2P, P2C, and E2V). Beyond conventional vertex-hyperedge interactions, we specifically introduce the C2P and P2C stages to facilitate information exchange across different hyperedge levels, as depicted in Figures \ref{fig:motivation}(d). Subsequently, we introduce a compact set of learnable query vectors to extract representative tabular structures from the encoder, optimized continuously through end-to-end fine-tuning. These vectors act as a Soft Structure Prompt \cite{lester2021power} prepended to the serialized text, naturally infusing the LLM with robust structural awareness. Remarkably, rather than relying on costly full-parameter tuning, H\textsuperscript{2}Table applies lightweight LoRA \cite{hu2022lora} to the base LLM, achieving performance comparable or even superior to full fine-tuning while incurring minimal computational overhead.
  
  To demonstrate that our structural modeling yields superior reasoning and generalization capabilities, we conduct extensive experiments on multiple complex table benchmarks. Our main contributions are summarized as follows:
\begin{itemize}
    \item \textbf{Hierarchical Nested Hypergraph Modeling for Complex Tables:} We propose a hierarchical nested hypergraph to preserve the intrinsic semantic entailment in multi-level tables. Furthermore, we develop a tailored, hierarchy-aware encoder that facilitates dynamic message passing to ensure comprehensive cross-level interactions.
    \item \textbf{Learnable Query Based Structural Alignment:} We introduce a query-based Soft Structure Prompt mechanism powered by a compact set of learnable query vectors. Through end-to-end fine-tuning, this mechanism progressively learns to extract task-beneficial structural features from tables, thereby boosting reasoning capabilities for question answering.
    \item \textbf{Comprehensive Superiority and Structural Robustness:} Empirical results show that H\textsuperscript{2}Table consistently outperforms competing baselines across multiple datasets. Crucially, it demonstrates exceptional robustness against increasing structural complexity, achieving an average relative improvement of 22.88\% on deep hierarchical tables and enabling fine-tuned medium-sized models to eclipse hundred-billion parameter models.
\end{itemize}

  \section{Related Work}
  \subsection{LLM-based Table Reasoning}

Currently, flattening two-dimensional tables into one-dimensional sequences is the prevailing approach for LLMs to process tabular data \citep{lu2025large}. To enhance the comprehension capabilities of LLMs on serialized tables, previous studies have typically employed instruction tuning (e.g., TableGPT, \citealp{li2024table}; TableLlama, \citealp{zhang2024tablellama}) or prompt engineering \citep[e.g., Chain-of-Table,][]{wang2024chain} for optimization. However, these optimization strategies exhibit distinct limitations: instruction tuning demands substantial amounts of data and incurs prohibitive computational costs while prompt engineering is heavily constrained by the context window limits. More importantly, forcibly flattening a table inherently disrupts its intrinsic topological structure, a critical flaw that becomes especially evident when dealing with complex or nested tables.

To mitigate the structural information loss caused by serialization, several studies (e.g., TabPrompt, \citealp{jin2023tabprompt}; HeGTa, \citealp{jin2025hegta}) have modeled tables as graphs to capture their intrinsic two-dimensional and higher-order features. However, since each edge in a conventional graph typically denotes only pairwise relationships, it struggles to accurately characterize the inherent one-to-many higher-order correlations between a row (or column) header and its constituent cells. Consequently, HYTREL \citep{chen2023hytrel} represents tables as hypergraphs, where a single hyperedge can connect multiple nodes. Although effective, it is restricted to simple flat tables and lacks integration with LLMs, making it incapable of handling generative downstream tasks such as TableQA. Building upon this, TAMO \citep{li2026table} integrates the HYTREL encoder with LLMs, treating tabular data as an independent continuous modality. While effective for flat tables, TAMO's encoding mechanism fundamentally flattens complex nested headers by directly grouping all associated cells under a single, coarse-grained hyperedge. This design inherently overlooks the multi-level dependencies and internal topological structures among the headers themselves. To bridge the gap,  H\textsuperscript{2}Table advances hypergraph-based table reasoning by explicitly modeling these hierarchical relationships. By designing nested hyperedges and a tailored multi-stage message-passing mechanism, our framework systematically captures fine-grained structural semantics while maintaining exceptional parameter efficiency.

\subsection{Cross-Modal Alignment between Graph Structures and Semantic Spaces}
The aforementioned graph-based tabular reasoning methods inherently rely on fusing the structural topology of graphs with the semantic comprehension capabilities of LLMs. Fundamentally, this constitutes a challenge of cross-modal feature alignment. Existing strategies generally fall into two paradigms. The first involves projection-based mapping, taking cues from large multimodal models (e.g., LLaVA, \citealp{liu2023visual}). For instance, GraphGPT \citep{tang2024graphgpt} employs an MLP to project node and subgraph embeddings, extracted by a pre-trained GNN, directly into the continuous input space of the LLM. Similarly, both HeGTa (utilizing heterogeneous graphs) and TAMO (leveraging hypergraphs) adopt analogous linear projectors for feature transformation. Alternatively, the second paradigm focuses on serialization-based alignment. InstructGLM \citep{ye2024language} heuristically flattens the graph's topological structure into an extended text sequence, facilitating subsequent fine-tuning strictly within the textual semantic domain.

However, simple linear projectors often struggle to efficiently compress highly complex structural information, while serialization inevitably leads to excessively long contexts and topological loss. In contrast to these conventional methods, we draw inspiration from the architecture of Q-Former \citep{li2023blip}. Specifically, we introduce learnable query vectors to perform cross-attention with graph embeddings, thereby actively extracting and compressing global topological and semantic features via end-to-end generative fine-tuning. This design not only alleviates the computational overhead but also significantly bolsters the model's feature aggregation capacity when reasoning over complex nested tables.

\begin{figure*}[t]
  \centering
  \includegraphics[width=0.95\textwidth]{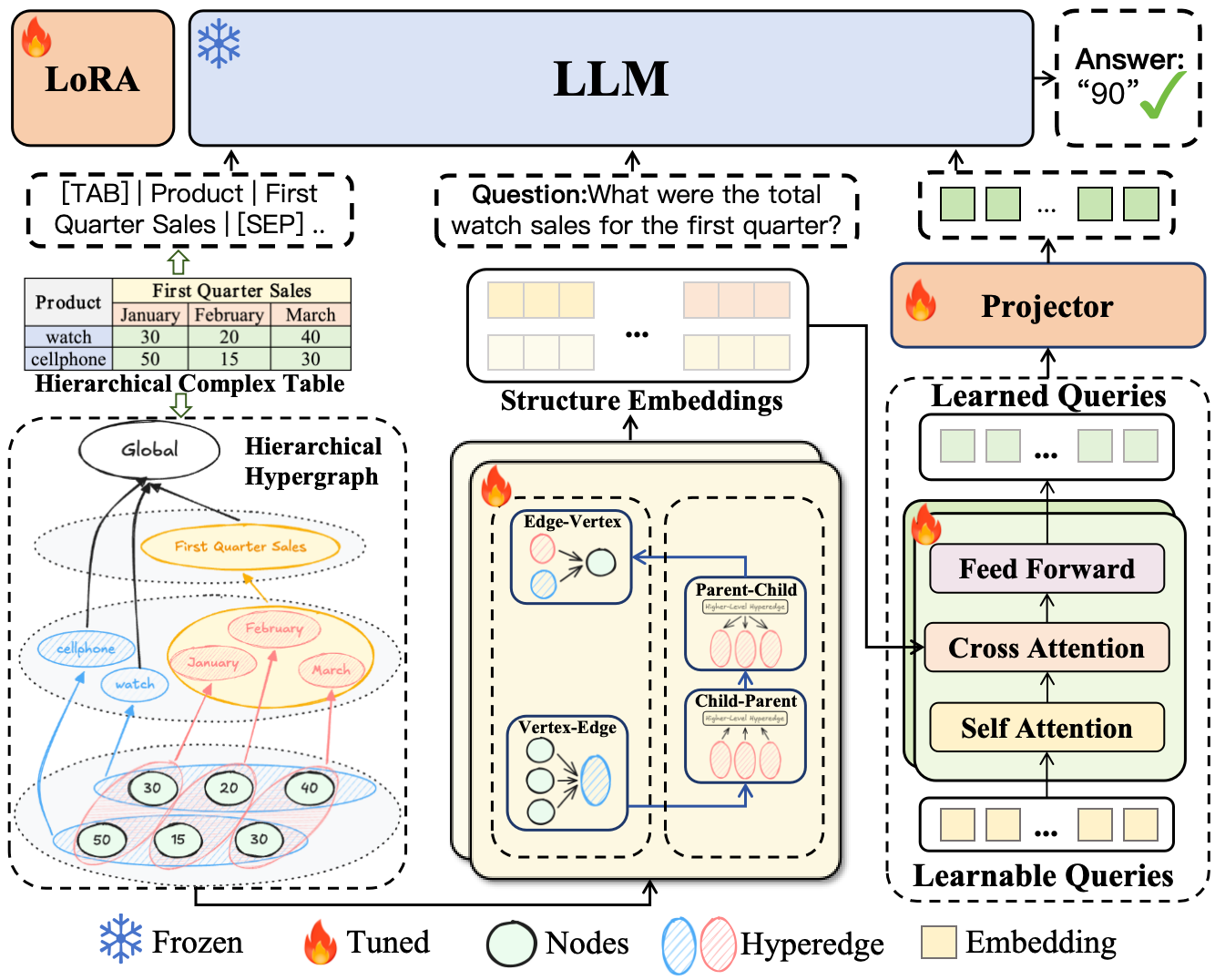}
  \caption{The overall architecture of the proposed framework. First, a complex hierarchical table is modeled as our designed hierarchical nested hypergraph, which is then processed by a hierarchical encoder featuring a four-stage message passing mechanism to derive the corresponding structural embeddings. Subsequently, a set of self-attended query vectors interacts with these structural embeddings via cross-attention. This yields a fixed number of structure-aware embeddings, which are finally combined with the serialized text and fed into the LLM to facilitate reasoning and generate the correct answer.}
  \label{fig:architecture}
\end{figure*}

  \section{Method}

  \subsection{Modeling Complex Tables as Hierarchical Nested Hypergraphs} \label{sec:hypergraph}

 To effectively capture the structural and semantic dependencies within complex tables, we model the table as a Hierarchical Nested Hypergraph $\mathcal{H}= (V, E, R, I)$. 
 
 Specifically, the node set $V = \{v_{1}, v_{2}, \dots, v_{n}\}$ represents all data cells. The hyperedge set $E = \{e_{1}, e_{2}, \dots, e_{m}\}$ comprises all table headers along with a virtual global root representing the entire table. This set is strictly partitioned into disjoint leaf headers $E_{leaf}$ and high-level headers $E_{high}$ (which contains the root). 
 
 The hierarchical relation $R \subseteq E \times E$ defines the structural dependency among hyperedges across different header levels, where a directed edge $(e_{i}, e_{j})$ indicates that $e_{i}$ is the parent of $e_{j}$. For instance, as shown in Figure \ref{fig:architecture}, in the complex table, the high-level header "First Quarter Sales" acts as the parent of the leaf header "March". The incidence relation $I \subseteq V \times E_{\text{leaf}}$ links each data cell $v$ to its corresponding lowest-level leaf headers. For instance, as illustrated in Figure \ref{fig:architecture}, the cell node "20" is connected to both the column header hyperedge "February" and the row header hyperedge "watch".
 
 Based on these relations, the hypergraph establishes a recursively nested scope. For any leaf header $e_{\text{child}} \in E_{\text{leaf}}$, its base scope is determined by the incidence relation as $S(e_{\text{child}}) = \{v \in V \mid (v, e_{\text{child}}) \in I\}$. Conversely, the scope of any high-level header $e_{\text{parent}} \in E_{\text{high}}$ is defined as the union of its children's scopes:$$S(e_{\text{parent}}) = \bigcup_{(e_{\text{parent}}, e_{\text{child}}) \in R} S(e_{\text{child}})$$Consequently, this formulation ensures that all parent hyperedges inherently perceive the scopes of their descending sub-headers all the way down to the individual cell nodes.

  \subsection{Hierarchy-Aware Hypergraph Encoder}
  Tailored to the proposed hierarchical nested hypergraph, we design a hierarchy-aware hypergraph encoder to facilitate comprehensive information propagation, encompassing both the interactions between data cells and hyperedges, and the hierarchical message passing among hyperedges across different levels. Within this encoder, we innovatively introduce a Hierarchical Hyperedge Interaction Module, which integrates a GAT into the hierarchical message-passing mechanism of hypergraphs. This design facilitates directed, weighted message propagation dynamically among hyperedges, thereby enabling the model to autonomously learn the underlying semantic correlations between table headers. As shown in Figure \ref{fig:architecture}, once a complex table is modeled as the hierarchical hypergraph detailed in Section \ref{sec:hypergraph} and initial embeddings are generated, these embeddings undergo iterative information interaction through this encoder. The encoder's single-layer forward propagation is meticulously structured into four distinct stages:
  \begin{itemize}
      \item \textbf{V2E (Vertex-to-Edge)}: Guided by the incidence relation $I$, this stage utilizes a set attention mechanism \citep{lee2019set, chen2023hytrel} to aggregate data cell features into their corresponding leaf headers, thereby establishing the base representations for the lowest-level hyperedges.
      \item \textbf{C2P (Child-to-Parent)}: Along the directed edges of the hierarchical relation $R$, a GAT dynamically weights and propagates information from child headers to their parents in a bottom-up manner. This enables high-level headers and ultimately the global root to capture the underlying data distribution and structural hierarchy.
      \item \textbf{P2C (Parent-to-Child)}: Operating symmetrically top-down along the reverse direction of $R$, this phase distributes global context. The GAT mechanism adaptively balances the broader semantic information passed from the parent with the preservation of the child's own local features.
      \item \textbf{E2V (Edge-to-Vertex)}: Finally, the hierarchically enriched hyperedge features are returned to the data cells via set attention. Residual connections are applied here to mitigate gradient vanishing, completing one full hierarchy-aware message-passing cycle.
  \end{itemize}
\subsection{Query-Based Parameter-Efficient Feature Alignment}
      While the hypergraph encoder captures rich topological representations, bridging the gap between continuous graph structural representations and the semantic reasoning space of LLMs remains a non-trivial challenge; naively mapping extensive graph embeddings via linear projection underutilizes structural priors and overloads the LLM's context. To address this, we introduce a parameter-efficient cross-modal alignment module to seamlessly bridge these distinct spaces.

    Inspired by Q-Former \cite{li2023blip}, we design a query-based mechanism for structural feature extraction and alignment. Specifically, we initialize a compact set of learnable query vectors designed to distill salient global topological representations from the dense node and hyperedge features yielded by the encoder. As illustrated in Figure \ref{fig:architecture}, the alignment module is instantiated via a stack of Transformer decoder layers \cite{vaswani2017attention}.

    In each layer, the query vectors first undergo self-attention to model their intra-query dependencies. During the subsequent cross-attention phase, these queries attend to the comprehensive structural embeddings acting as keys and values generated by the hypergraph encoder, thereby facilitating deep cross-modal fusion \cite{li2023blip}. Finally, the fused representations are projected through a feed-forward network (FFN).

    Formally, let $H$ denote the structural graph features and $Q^{(l-1)}$ denote the queries at layer $l-1$; the core update process at the $l$-th layer can be abstracted as:$$ Q^{(l)} = \text{FFN}\Big( \text{MCA}\big( \text{MSA}(Q^{(l-1)}), H, H \big) \Big) $$where $\text{MSA}$ and $\text{MCA}$ represent multi-head self-attention and cross-attention, respectively. For brevity, the standard residual connections and layer normalization applied within each sub-module are omitted from the notation.

    Optimized continuously through end-to-end fine-tuning, the queries learn to extract the most representative structural features of the table. After multi-layer alignment, these highly distilled query embeddings function as soft structural prompts that are prepended to the serialized table text, constructing a cohesive dual-stream semantic-structural input that seamlessly guides the downstream LLM's reasoning \cite{liu2023pre}.

    \begin{table*}[t]
    \centering
    % 使用 resizebox 将表格自动缩放至与正文等宽
    \resizebox{\textwidth}{!}{% 
    \begin{tabular}{l ccccc cccc}
        \toprule
        \textbf{Dataset} & \multicolumn{5}{c}{\textbf{HiTab}} & \multicolumn{4}{c}{\textbf{TATQA}} \\
        \cmidrule(lr){2-6} \cmidrule(lr){7-10}
        \textbf{Depth} & I & II & III & IV & \textbf{Avg.} & I & II & III & \textbf{Avg.} \\
        \midrule
        
        \textbf{Llama3.1-8B} & 0.3143 & 0.3609 & 0.2977 & 0.2286 & 0.3004 & 0.1866 & 0.1693 & 0.1603 & 0.1721 \\
        \hspace{1em}+pure text (LoRA) & 0.6571 & 0.7190 & 0.6458 & 0.5714 & 0.6483 & 0.3528 & \underline{0.3964} & 0.3462 & 0.3651 \\
        \hspace{1em}+TAMO (LoRA)       & \textbf{0.7143} & \underline{0.7960} & \underline{0.7355} & \underline{0.6286} & \underline{0.7186} & \underline{0.4461} & \textbf{0.4819} & \underline{0.4295} & \underline{0.4525} \\
        \hspace{1em}+\textbf{H\textsuperscript{2}Table (LoRA)} & \underline{0.6857} & \textbf{0.8046} & \textbf{0.7614} & \textbf{0.7143} & \textbf{0.7415} & \textbf{0.4606} & \textbf{0.4819} & \textbf{0.4359} & \textbf{0.4595} \\
        \midrule
        
        \textbf{Gemma2-9B}   & \underline{0.6000} & 0.6733 & 0.5818 & 0.5143 & 0.5924 & 0.3965 & 0.4136 & 0.3974 & 0.4025 \\
        \hspace{1em}+pure text (LoRA) & \textbf{0.7143} & 0.7946 & 0.7343 & \underline{0.7143} & \underline{0.7394} & \underline{0.4781} & \underline{0.5328} & \textbf{0.5128} & \underline{0.5079} \\
        \hspace{1em}+TAMO (LoRA)       & \textbf{0.7143} & \underline{0.8031} & \underline{0.7540} & 0.5714 & 0.7107 & 0.4723 & 0.5311 & 0.4936 & 0.4990 \\
        \hspace{1em}+\textbf{H\textsuperscript{2}Table (LoRA)} & \textbf{0.7143} & \textbf{0.8260} & \textbf{0.7552} & \textbf{0.7429} & \textbf{0.7596} & \textbf{0.4927} & \textbf{0.5354} & \underline{0.5000} & \textbf{0.5094} \\
        \midrule
        
        \textbf{Llama2-7B}   & 0.2000 & 0.1997 & 0.1242 & 0.1429 & 0.1667 & 0.1283 & 0.1364 & 0.1410 & 0.1352 \\
        \hspace{1em}+pure text (LoRA) & 0.5429 & \underline{0.5649} & 0.4379 & 0.4000 & 0.4864 & 0.1953 & 0.2409 & 0.1859 & 0.2074 \\
        \hspace{1em}+TAMO (LoRA)       & \underline{0.6000} & \textbf{0.6862} & \underline{0.5461} & \underline{0.4571} & \underline{0.5724} & \underline{0.3324} & \textbf{0.3575} & \textbf{0.3718} & \underline{0.3539} \\
        \hspace{1em}+\textbf{H\textsuperscript{2}Table (LoRA)} & \textbf{0.6571} & \textbf{0.6862} & \textbf{0.5720} & \textbf{0.5714} & \textbf{0.6217} & \textbf{0.3615} & \underline{0.3549} & \underline{0.3654} & \textbf{0.3606} \\
        \midrule
        
        TableLlama & 0.5714 & 0.6933 & 0.5966 & 0.5429 & 0.6011 & 0.1574 & 0.1451 & 0.1282 & 0.1436 \\
        GPT-4o     & 0.7143 & 0.7418 & 0.6913 & 0.6571 & 0.7011 & 0.5335 & 0.5337 & 0.5064 & 0.5245 \\
        DeepseekV3 & 0.6571 & 0.7247 & 0.6531 & 0.6286 & 0.6659 & 0.4606 & 0.4491 & 0.4551 & 0.4549 \\
        
        \bottomrule
    \end{tabular}%
    }
    \caption{Performance comparison of H\textsuperscript{2}Table against baseline models on the TableQA task across different table nesting depths (I--IV for HiTab, and I--III for TATQA). Models without indentation (including backbone baselines and the bottom group) are evaluated under the zero-shot setting, while indented rows represent LoRA fine-tuned variants. ``Avg.'' denotes the arithmetic mean across depths. \textbf{Bold} and \underline{underlined} values indicate the best and second-best results within each model group, respectively.}
    \label{tab:main_results}
\end{table*}

     Irrespective of table complexity, the module condenses structural information into a fixed number of tokens, alleviating the sequence explosion common in direct graph inputs and minimizing the LLM's contextual burden. Moreover, as this module manages the cross-modal translation, it circumvents costly full-parameter fine-tuning. By updating merely 1\% of the parameters, this paradigm achieves performance comparable to full fine-tuning. Consequently, it drastically reduces computational costs while mitigating catastrophic forgetting and preserving the LLM’s inherent reasoning capabilities.

  \section{Experiments}

  \subsection{Experimental setup}

\begin{figure}[t]
      \centering
      \includegraphics[width=0.48\textwidth]{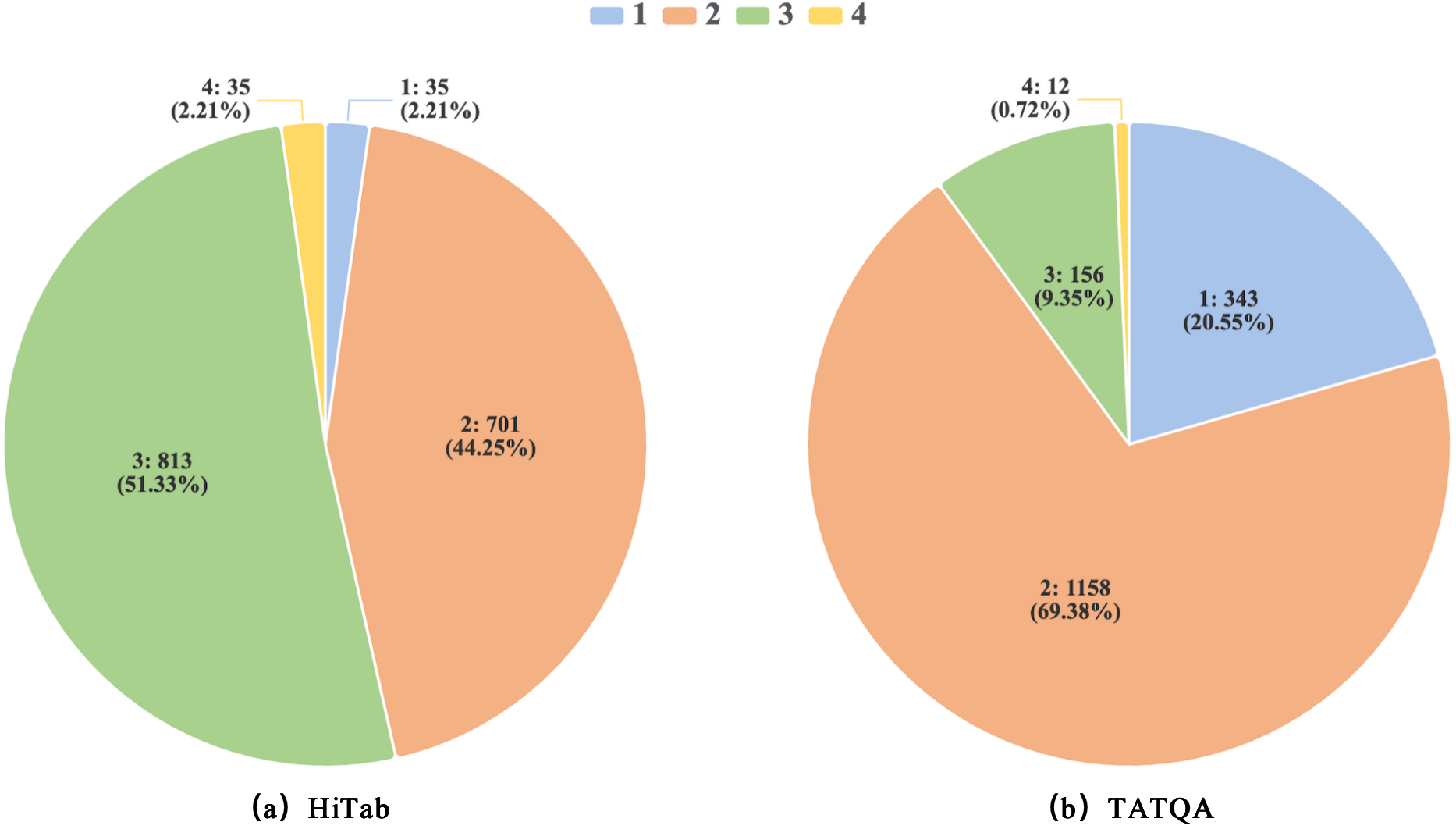}
      \caption{Distribution of table nesting depth for HiTab and TATQA}
      \label{fig:data_distribution}
    \end{figure}

    \paragraph{Datasets.} We evaluate H\textsuperscript{2}Table on several representative benchmarks featuring complex hierarchical structures, namely HiTab \cite{cheng2022hitab} and TATQA \cite{zhu2021tat}. The primary downstream task across all selected datasets is TableQA, with Accuracy serving as the core evaluation metric. The distribution of tables within these datasets, categorized by header nesting depth, is illustrated in Figure \ref{fig:data_distribution}. Notably, because the formatted TATQA dataset includes a mere 12 tables with four-level headers, a quantity insufficient for statistical significance, these instances were excluded from our main experimental evaluation.

    \paragraph{Baselines.} We select TableLlama \cite{zhang2024tablellama} as a primary baseline, as it achieved state-of-the-art performance across numerous table reasoning tasks through extensive SFT on diverse tabular datasets. Additionally, we compare our approach with TAMO \cite{li2026table}, a contemporary method that treats tables as an independent modality and has demonstrated superior performance over TableLlama on several benchmarks. To ensure a fair and direct comparison with these representative models, we adopt Llama2-7B \cite{touvron2023llama} as our primary base model. Furthermore, to evaluate the efficacy of our method on more advanced architectures, we extend our experiments to Llama3.1-8B \cite{grattafiori2024llama} and Gemma2-9B \cite{team2024gemma}. Our evaluation also includes comparisons against several training configurations for the base models, including zero-shot inference, and LoRA fine tuning using pure-text representations. Detailed experimental results are reported in Table \ref{tab:main_results}.

    \paragraph{Implementation Details}
    For the initialization of our hypergraph, node embeddings are extracted from the final layer of a pre-trained RoBERTa-large model \cite{liu2019roberta}, and the hidden size of our hypergraph encoder is strictly set to 1024 to maintain dimensional consistency. For a fair comparison, both the baseline models and our proposed framework are fine-tuned end-to-end using LoRA with a rank of 8, with binary cross-entropy as the loss function. More detailed hyperparameter settings and training configurations are provided in Appendix~\ref{sec:experiments}.

  \subsection{Main Results}
% Please ensure your table has the label \label{tab:main_results} or change the reference below.
Table~\ref{tab:main_results} presents the performance comparison on the HiTab and TATQA datasets. To evaluate the capability of handling complex tables, the results are further categorized by table nesting depth. Overall, our proposed H\textsuperscript{2}Table consistently outperforms all baseline representations across different base models and depth levels, demonstrating its effectiveness in TableQA tasks especially when encountering hierarchical complex tables .

When integrated with various open-source LLMs such as Llama3.1-8B, Gemma2-9B, and Llama2-7B, the H\textsuperscript{2}Table representation consistently yields the highest average accuracy within each base model group. For instance, fine-tuning Gemma2-9B with H\textsuperscript{2}Table achieves an average accuracy of 0.7596 on HiTab, substantially surpassing both the pure text and TAMO representations. This indicates that H\textsuperscript{2}Table serves as a general and robust paradigm to improve the table reasoning capabilities of LLMs.

Aligning with our primary motivation to tackle complex tables, H\textsuperscript{2}Table exhibits remarkable superiority on tables with high nesting depths (i.e., Depth-3 and Depth-4). While the performance of traditional baseline methods drops precipitously as structural complexity increases, H\textsuperscript{2}Table remains highly robust. On the HiTab dataset at Depth-4, Llama3.1-8B with H\textsuperscript{2}Table achieves an accuracy of 0.7143, yielding a significant absolute improvement over the pure text (0.5714) and TAMO (0.6286) baselines. Similar trends are consistently observed across other base models, validating that our method effectively preserves structural integrity and captures deep hierarchical semantics.

Remarkably, fine-tuning medium-sized models (7B--9B parameters) with H\textsuperscript{2}Table enables them to rival or even surpass state-of-the-art closed-source and massively large LLMs in complex table reasoning.  On the HiTab dataset, by only updating approximately 1\% of the total parameters, Gemma2-9B equipped with H\textsuperscript{2}Table (0.7596) significantly outperforms the hundred-billion parameter DeepseekV3 (0.6659), and the powerful GPT-4o (0.7011). These results compellingly demonstrate that providing a superior structural representation empowers smaller models to achieve highly competitive reasoning capabilities on complex tables.

\subsection{Ablation Study}
To thoroughly investigate the individual contributions and architectural synergies of key components in H\textsuperscript{2}Table, we conduct an extensive ablation study on the HiTab dataset based on Llama3.1-8B.

\begin{figure}[H]
      \centering
      \includegraphics[width=0.48\textwidth]{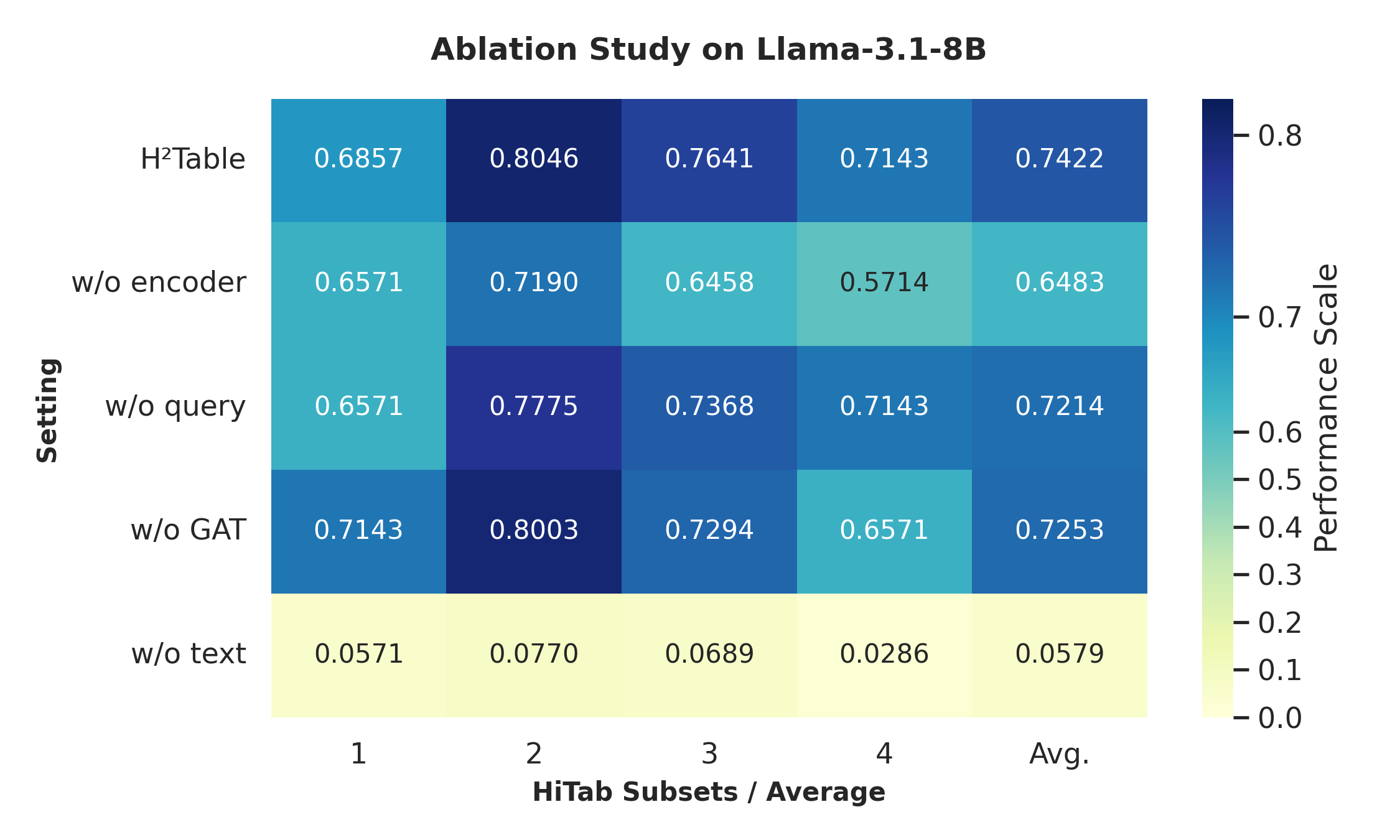}
      \caption{The heatmap of the ablation study. Deep blue hues indicate superior performance, while lighter shades represent degraded capabilities.}
      \label{fig:ablation}
\end{figure}

\begin{table*}[t]
    \centering
    % l 代表第一列左对齐，c 代表后续数据列居中对齐
    \begin{tabular}{l ccc ccc}
        \toprule
        \textbf{Training Set} & \multicolumn{3}{c}{HiTab} & \multicolumn{3}{c}{TATQA} \\
        \cmidrule(lr){2-4} \cmidrule(lr){5-7}
        \textbf{Test Set} & HiTab & TATQA & AITQA & TATQA & HiTab & AITQA \\
        \midrule
        
        \textbf{Llama3.1-8B} & 0.3245 & 0.1708 & 0.5146 & - & - & - \\
        \hspace{1em}+Prompt tuning   & 0.0322 & 0.0298 & 0.1023 & 0.0639 & 0.0463 & 0.0498 \\
        \hspace{1em}+pure text (LoRA)& 0.6774 & 0.2548 & 0.8330  & 0.3815 & 0.5696 & 0.7540 \\
        \hspace{1em}+TAMO (LoRA)      & \underline{0.7643} & \underline{0.3156} & \underline{0.8634} & \underline{0.4721} & \underline{0.6086} & \underline{0.7974}  \\
        \hspace{1em}+\textbf{H\textsuperscript{2}Table (LoRA)} & \textbf{0.7713} & \textbf{0.3225} & \textbf{0.8673} & \textbf{0.4745} & \textbf{0.6313} & \textbf{0.8220} \\
        \bottomrule
    \end{tabular}
    \caption{Performance comparison of various models on standard and cross-domain TableQA tasks. \textbf{Bold} and \underline{underlined} values indicate the best and second-best performance among the fine-tuned variants for each base model, respectively. All results are averaged over three independent training and evaluation runs.}
    \label{tab:generalization}
\end{table*}

As visualized in Figure~\ref{fig:ablation}, removing the textual feature module ($\text{w/o text}$) causes a catastrophic performance collapse (represented by the blank row), underscoring that textual semantics remain the foundation of table understanding. Crucially, omitting the complete hierarchical hypergraph encoder ($\text{w/o encoder}$) also induces severe performance degradation, which validates that explicit structural modeling is indispensable for breaking the table-comprehension bottleneck of plain LLMs.

Furthermore, removing the GAT-based message passing (denoted as $\text{w/o GAT}$, where the hierarchical edge aggregation is replaced by a standard Set Attention mechanism) or omitting the cross-attention alignment ($\text{w/o query}$) leads to a slight yet consistent performance decline across most subsets. Although these performance gaps may appear modest, such stable improvements are highly meaningful given the extremely challenging generative setting of complex hierarchical TableQA. 

Interestingly, the $\text{w/o GAT}$ variant yields highly competitive results and even achieves the top score on the Depth-1 subset. As Depth-1 tables are essentially flat, they do not heavily rely on complex hierarchical message passing; thus, the strong performance of this variant is expected. However, the performance gap becomes significantly more pronounced on deeper hierarchical tables (e.g., Depth-3 and Depth-4), which is exactly where the GAT-based hierarchy-aware propagation demonstrates its efficacy. Overall, the full H\textsuperscript{2}Table architecture achieves the best average performance, validating the necessity and strong synergistic effects of our integrated modules.

% \begin{table}[H]
% \centering
% \setlength{\tabcolsep}{5pt} % 稍微收窄列间距，使缩放后的字体更大、更美观
% \resizebox{\columnwidth}{!}{% % 自动缩放到单栏宽度
% \begin{tabular}{lccccc}
% \toprule
% \multirow{2}{*}{\textbf{Setting}} & \multicolumn{5}{c}{\textbf{HiTab}} \\
% \cmidrule{2-6}
%  & \textbf{1} & \textbf{2} & \textbf{3} & \textbf{4} & \textbf{Avg.} \\
% \midrule
% H\textsuperscript{2}Table & \underline{0.6857} & \textbf{0.8046} & \textbf{0.7641} & \textbf{0.7143} & \textbf{0.7422} \\
% w/o encoder & 0.6571 & 0.7190 & 0.6458 & 0.5714 & 0.6483 \\
% w/o query & 0.6571 & 0.7775 & \underline{0.7368} & \textbf{0.7143} & 0.7214 \\
% w/o GAT & \textbf{0.7143} & \underline{0.8003} & 0.7294 & \underline{0.6571} & \underline{0.7253} \\
% w/o text & 0.0571 & 0.0770 & 0.0689 & 0.0286 & 0.0579 \\
% \bottomrule
% \end{tabular}%
% }
% \caption{Ablation study on HiTab based on Llama3.1-8B. Bold and underlined values indicate the best and second-best performance, respectively.}
% \label{tab:ablation}
% \end{table}

\subsection{Out-of-Distribution Generalization}
To evaluate the out-of-distribution (OOD) generalization capability of our method, we conduct cross-domain experiments by incorporating AITQA \cite{katsis2022ait}, another complex hierarchical table benchmark, as an unseen evaluation set. 

As shown in Table~\ref{tab:generalization}, the vanilla Llama3.1-8B and its prompt-tuning variant exhibit poor robustness across domains, with prompt tuning suffering from catastrophic performance drops. Incorporating table-specific structural modeling significantly boosts generalization. For instance, when trained on HiTab, the pure text baseline (LoRA) achieves an OOD score of 0.8330 on AITQA, whereas the structure-aware TAMO improves it to 0.8634.

Most importantly, our proposed H\textsuperscript{2}Table consistently achieves the best performance across all settings, outperforming all baselines by a clear margin. Specifically, whether trained on HiTab or TATQA, H\textsuperscript{2}Table uniformly surpasses TAMO on both in-domain and all cross-domain test sets. These results demonstrate that by effectively encoding hierarchical table structures, H\textsuperscript{2}Table learns more robust, domain-invariant representations for complex TableQA tasks.

\subsection{Analysis on Number of Query Vectors}

As shown in Table \ref{tab:query_vectors}, we evaluated the impact of varying the number of query vectors on overall performance. 

\begin{table}[htbp]
  \centering
  \begin{tabular}{lccc}
    \toprule
    N & Llama2-7B & Llama3.1-8B & Gemma2-9B \\
    \midrule
    1  & \textbf{0.6172} & 0.7675 & \underline{0.7862} \\
    4  & 0.5905 & \underline{0.7677} & 0.7847 \\
    8  & 0.6061 & 0.7672 & 0.7841 \\
    12 & \underline{0.6113} & \textbf{0.7713} & 0.7860 \\
    16 & 0.5995 & 0.7603 & \textbf{0.7914} \\
    \bottomrule
  \end{tabular}
  \caption{Comparison of different numbers of query vectors (denoted as N). Bold and underlined values indicate the best and second-best performance, respectively.}
  \label{tab:query_vectors}
\end{table}

While the optimal vector count differs slightly across model architectures, using 12 query vectors consistently yields robust and highly competitive results. Specifically, it achieves the best performance on Llama3.1-8B and maintains near-optimal scores for both Llama2-7B and Gemma2-9B. To strike a balance between performance and cross-model stability, we empirically set the number of query vectors to 12 for all experiments.

  \section{Conclusion}

  In this paper, we propose a novel framework, H\textsuperscript{2}Table. Unlike conventional methods that model tables as standard hypergraphs, we design a specialized hierarchical hypergraph representation tailored for complex tables, coupled with a hypergraph encoder featuring a four-stage hierarchical message-passing mechanism. Subsequently, a learnable query vector mechanism is employed to extract the most representative structural features from the encoder's output. The resulting structural embeddings effectively assist LLMs in table reasoning tasks. The effectiveness of our framework has been validated across multiple datasets, particularly those featuring intricate hierarchical tables, demonstrating strong performance and generalization capabilities.

  \section*{Limitations}

Our current work assumes that the header structures of complex hierarchical tables can be accurately obtained beforehand. Under this setting, we treat header hierarchies as available structural annotations to isolate and evaluate the contribution of hierarchical hypergraph modeling itself. Future work could explore integrating upstream table-structure extractors with H\textsuperscript{2}Table for end-to-end processing. Additionally, the proposed framework still relies on feeding the serialized tabular text into the LLM. Subsequent work will investigate how to enable the structural embeddings generated by the encoder to simultaneously capture rich semantic information.

  \section*{Acknowledgments}
  We gratefully acknowledge the support of the Key Research and Development Program of Heilongjiang Province, China [2024ZX01A07] and the National Key Research and Development Program [2025YFE0209200].


  \bibliography{custom}

  \appendix

  \section{Details of the Experiments}
  \label{sec:experiments}
  
\begin{figure*}[t]
  \centering
  \includegraphics[width=\textwidth]{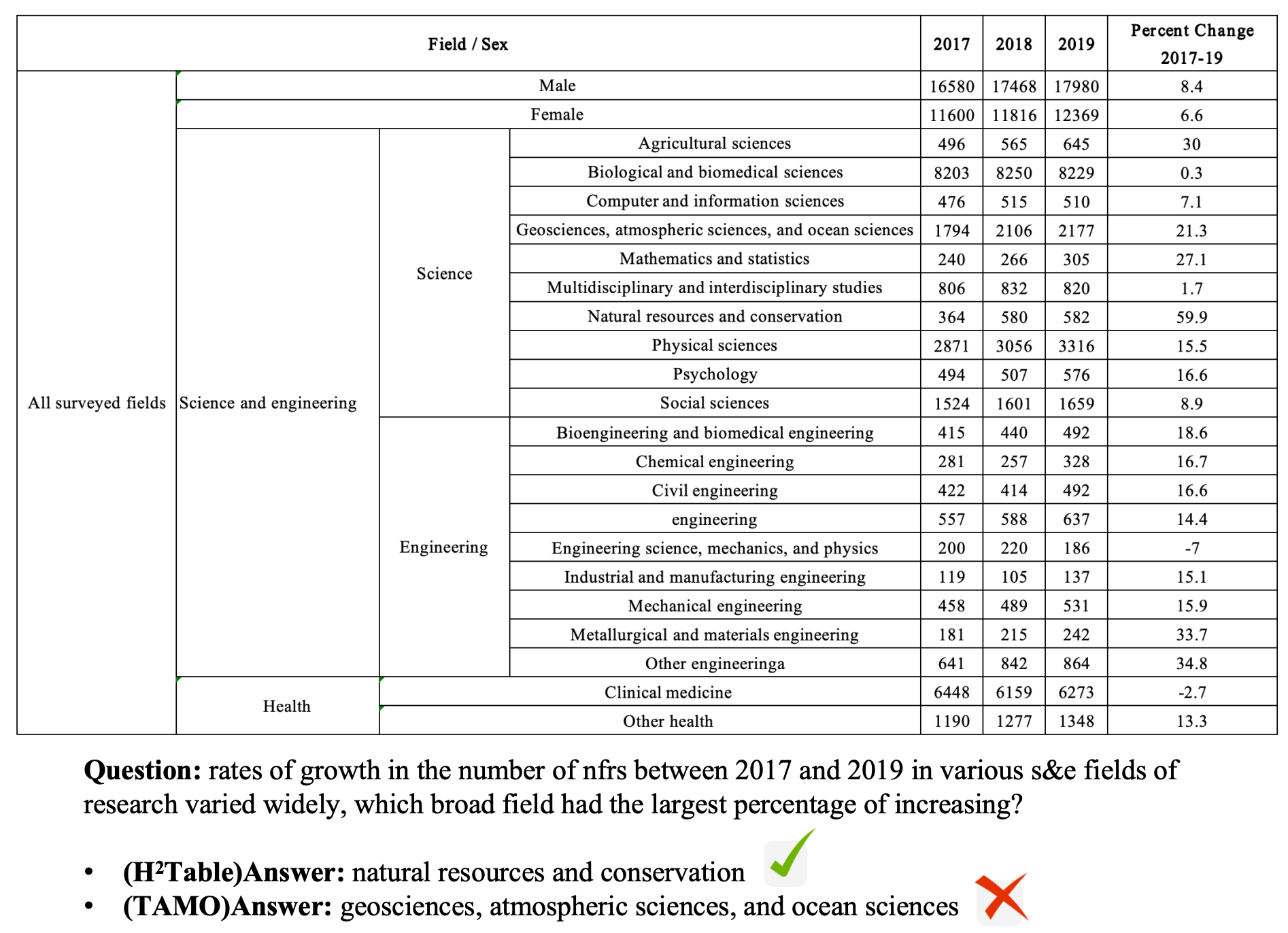}
  \caption{An example of complex hierarchical tables which has a depth of 4 and corresponding question.}
  \label{fig:case}
\end{figure*}

  \subsection{Training Settings.}  
  For efficient fine-tuning, we applied LoRA ($r=8, \alpha=16$, dropout $= 0.05$). The model was optimized using AdamW ($\beta_1=0.9, \beta_2=0.95$, weight decay $= 0.05$) with a learning rate of $1 \times 10^{-5}$.To construct the training sequences, we concatenated the graph/query embeddings with the text prompt (comprising the description and the question) and the target label. Descriptions were truncated to 1,024 tokens, and the target generation length was bounded to 128 tokens. While the training sequences were dynamically batched, we relaxed the maximum context window to 4,096 tokens during the inference and evaluation phases.

  \subsection{Datasets.} In addition to HiTab, we applied specific preprocessing steps to the other two datasets. For AITQA, since the original dataset provides annotations for all parent headers of each leaf header, it was straightforward to convert it into the HiTab format, specifically, comprising a row header tree, a column header tree, and a data matrix. For TATQA, we employed LLM to transform the original flat matrix tables into the aforementioned hierarchical format. Consequently, this approach inevitably introduces some conversion errors, which partially explains why the model trained on the TATQA dataset exhibits less stable performance compared to the model trained on HiTab. A detailed analysis of this noise is provided below.

While converting flat TATQA tables into hierarchical structures via LLMs inherently introduces a degree of structural noise, all evaluated models (both baselines and H\textsuperscript{2}Table) are trained and tested on this identical augmented dataset. Consequently, the relative performance comparisons remain strictly fair.

To quantitatively assess the impact of this conversion noise, we manually reviewed a random sample of 100 tables from the TATQA test set (comprising 600 QA pairs, averaging 6 questions per table). The manual inspection revealed that 91 tables were converted accurately, whereas 9 exhibited noticeable structural errors. This low error rate (9\%) suggests that TATQA tables possess predominantly simple, shallow hierarchies, making the LLM-based structural extraction largely reliable for this benchmark. 

To further investigate how this structural noise affects downstream reasoning, we evaluated H\textsuperscript{2}Table (using the Gemma-2-9B backbone) on this sampled subset. The results reveal a clear performance disparity between correctly and incorrectly converted tables:

\begin{table}[htbp]
  \centering
  \begin{tabular}{lc}
    \toprule
    \textbf{Subset} & \textbf{Accuracy} \\
    \midrule
    \textbf{Correct} (546 QA pairs) & \textbf{52.20\%} \\
    \textbf{Wrong} (54 QA pairs) & \textbf{37.04\%} \\
    \midrule
    \textbf{Overall} (Full TATQA Test Set) & \textbf{52.07\%} \\
    \bottomrule
  \end{tabular}
  \caption{Performance disparity between correctly and incorrectly converted tables. The overall evaluation encompasses the full TATQA test set of 1,669 QA pairs.}
  \label{tab:noise_analysis}
\end{table}

As detailed above, the model achieves a 52.20\% accuracy on the correctly converted subset, which is 15.16\% higher than its performance on the subset with structural errors (37.04\%). This discrepancy confirms that structural conversion quality directly influences downstream TableQA performance. However, because the overall conversion error rate remains low, the accuracy on the full TATQA test set (52.07\%) aligns closely with the performance on the accurately converted subset, thereby demonstrating the robustness of our evaluation on this benchmark.

  \subsection{Evaluation.} Regarding the evaluation metrics, rather than employing a strict Exact Match (EM), we applied necessary data cleaning and format normalization to the model's predictions, such as stripping currency symbols (e.g., \$). We clarify that all compared methods (including closed-source models, fine-tuned baselines, and H\textsuperscript{2}Table) are evaluated under exactly the same answer-cleaning and format-normalization criteria. The reported accuracy is therefore computed under a unified relaxed-match protocol after shared post-processing, ensuring fair comparability. 

  \section{Case Study}
  \label{sec:case}

In this section, we present a case study to demonstrate our model's capability in handling complex table structures. As illustrated in Figure \ref{fig:case}, our model successfully yields the correct answer to the given question, whereas a strong baseline, TAMO, fails to do so. This superior performance is attributed to our specifically designed modeling and hierarchical message-passing mechanism, which enable us to process such intricate tables more effectively.

\section{Discussion}
\subsection{Compatibility and Generalization over Flat Hypergraphs}
A natural theoretical question is how H\textsuperscript{2}Table behaves when applied to standard flat tables lacking nested structures. In such degenerate cases, where all headers reside at a single depth, the hierarchical relations naturally collapse, and message passing reduces strictly to the leaf-level vertex-hyperedge interactions (i.e., V2E and E2V).

We regard this property as an intrinsic strength of our design: H\textsuperscript{2}Table remains mathematically well-defined and seamlessly applicable to flat tables, without imposing any artificial or redundant hierarchical priors. Crucially, even under this flat degeneration, H\textsuperscript{2}Table is fundamentally distinct from prior hypergraph-based approaches such as TAMO in how structural knowledge is transferred to the LLM. Rather than relying on simple average-pooling over encoder embeddings, our query-based cross-attention module employs learnable queries to selectively attend over encoder representations, effectively distilling high-density structural signals into a compact soft prompt. Furthermore, once multi-level nested headers are present, the hierarchical message-passing stages (C2P and P2C) are dynamically activated, empowering the model to capture cross-level semantic entailments that flat hypergraph formulations fundamentally overlook. Consequently, H\textsuperscript{2}Table generalizes rather than merely replaces flat hypergraph modeling---it preserves full backward compatibility with flat tables, provides a more selective encoder-LLM alignment mechanism, and explicitly incorporates hierarchical inductive bias precisely when nested headers matter.

\subsection{Statistical Significance Analysis on Deep Hierarchy Tables}
To ensure a robust statistical evaluation and address the limited sample size of the isolated Depth-4 subset ($N=35$), we aggregate the Depth-3 ($N=813$) and Depth-4 tables into a unified "Deep Hierarchy" subset, yielding a more substantial total sample size of $N=848$. Evaluated using the Llama-3.1-8B backbone, our proposed H\textsuperscript{2}Table architecture demonstrates a stable improvement over the strongest baseline, TAMO, on this combined challenging subset.

To formally validate the reliability of these performance gains, we conduct a paired McNemar's test comparing the exact match predictions of H\textsuperscript{2}Table against TAMO. The test confirms that the improvements introduced by our model are statistically significant, yielding an exact McNemar $p$-value of 0.0079 and a continuity-corrected $\chi^2$ $p$-value of 0.0083 ($p < 0.01$). These statistical results substantiate that the advantages of H\textsuperscript{2}Table on highly complex hierarchical tables are robust and not an artifact of small sample variance.

\end{document}